\documentclass[conference]{IEEEtran}
\usepackage{cite}
\usepackage{textcomp}
\usepackage{dblfloatfix}
\usepackage{amsmath,amssymb,amsfonts}
\usepackage{algorithmic}
\usepackage{graphicx}
\usepackage{textcomp}
\usepackage{footnote}
\usepackage{booktabs}
\usepackage{graphicx}
\usepackage{multirow}
\usepackage{subfig}
\usepackage{array}
\usepackage{mathtext}
\usepackage{enumitem}
\usepackage{amsmath}
\usepackage{algorithmic}
\usepackage{algorithm}
\usepackage{epsfig}
\usepackage{url}
\usepackage{epstopdf}
\usepackage{cite}
\usepackage{setspace}
\usepackage{xcolor}
\usepackage{xcolor}
\usepackage{caption}
\usepackage{tabularray}
\UseTblrLibrary{booktabs}

\renewcommand{\v}[1]{{\mathbf{#1}}}

\newcommand\MyBox[2]{
  \fbox{\lower0.75cm
    \vbox to 1.7cm{\vfil
      \hbox to 1.7cm{\hfil\parbox{1.4cm}{#1\\#2}\hfil}
      \vfil}%
  }%
}

\usepackage{xcolor}
\usepackage{subfig}

\makeatletter
\def\ps@IEEEtitlepagestyle{
  \def\@oddfoot{\mycopyrightnotice}
  \def\@evenfoot{}
}
\def\mycopyrightnotice{
  {\footnotesize
  \begin{minipage}{\textwidth}
  \centering
  Copyright~\copyright~2022 IEEE. Personal use of this material is permitted. Permission from IEEE must be obtained for all other uses, in any current or future media, including reprinting/republishing this material for advertising or promotional purposes, creating new collective works, for resale or redistribution to servers or lists, or reuse of any copyrighted component of this work in other works.
  \end{minipage}
  }
}
\newcolumntype{f}{>{$}l<{$}}
\newcolumntype{n}{l}
\newcolumntype{N}{>{\normal}c}
\newcolumntype{v}[1]{>{\centering\hspace{0pt}}p{#1}}
\newcolumntype{V}[1]{>{\scriptsize\raggedright\hspace{2pt}}P}

\newcolumntype{B}[1]{>{\boldmath\DC@{.}{,}{#1}}l<{\DC@end}}
\newcolumntype{d}[1]{>{\DC@{.}{,}{#1}}l<{\DC@end}}
\newcolumntype{i}[1]{>{\DC@{.}{,}{#1}\mathnormal\bgroup}l<{\egroup\DC@end}}
\newcolumntype{s}[1]{>{\DC@{.}{,}{#1}\mathsf\bgroup}l<{\egroup\DC@end}}

\def\BibTeX{{\rm B\kern-.05em{\sc i\kern-.025em b}\kern-.08em
    T\kern-.1667em\lower.7ex\hbox{E}\kern-.125emX}}

\begin{document}
\bibliographystyle{ieeetr}
\title{Building Matters: Spatial Variability in Machine Learning Based Thermal Comfort Prediction in Winters}
\author{\IEEEauthorblockN{Betty Lala$^\Phi$, Srikant Manas Kala$^\dag$, Anmol Rastogi$^*$, Kunal Dahiya$^\P$, Hirozumi Yamaguchi$^\dag$, Aya Hagishima$^\Phi$}
\IEEEauthorblockA{$^\Phi$ Interdisciplinary Graduate School of Engineering Sciences, Kyushu University, Fukuoka, Japan\\ $^\dag$ Graduate School of Information Science and Technology, Osaka University, Japan\\ $^*$ Indian Institute of Technology Hyderabad, India $^\P$ Indian Institute of Technology Delhi, India\\
Email: lala.betty.919@s.kyushu-u.ac.jp, manas\_kala@ist.osaka-u.ac.jp, ai19btech11021@iith.ac.in, kunalsdahiya@gmail.com\\h-yamagu@ist.osaka-u.ac.jp, ayahagishima@kyudai.jp}}

\maketitle
\begin{abstract}
Thermal comfort in indoor environments has an enormous impact on the health, well-being, and performance of occupants.
Given the focus on energy efficiency and Internet of Things enabled smart buildings, machine learning (ML) is being increasingly used for data-driven thermal comfort (TC) prediction. Generally, ML-based solutions are proposed for air-conditioned or HVAC ventilated buildings and the models are primarily designed for adults. 
On the other hand, naturally ventilated (NV) buildings are the norm in most countries. They are also ideal for energy conservation and long-term sustainability goals.
However, the indoor environment of NV buildings lacks thermal regulation and varies significantly across spatial contexts. These factors make TC prediction extremely challenging.
Thus, determining the impact of building environment on the performance of TC models is important. Further, the generalization capability of TC prediction models across different NV indoor spaces needs to be studied.
This work addresses these problems. 
Data is gathered through month-long field experiments conducted in 5 naturally ventilated school buildings, involving 512 primary school students. 
The impact of spatial variability on student comfort is demonstrated through variation in prediction accuracy (by as much as 71\%). The influence of building environment on TC prediction is also demonstrated through variation in feature importance. Further, a comparative analysis of spatial variability in model performance is done for children (our dataset) and adults (ASHRAE-II database). Finally, the generalization capability of thermal comfort models in NV classrooms is assessed and major challenges are highlighted. 
\end{abstract}




Thermal Comfort, Machine Learning, Spatial Variability, Natural Ventilation, Energy Efficiency, IoT, Sensors, Students, Classrooms, Multi-class Classification 


\section{Introduction}
\label{sec:sample1}
We spend over 90\% of our time in indoor spaces \cite{[2]}. As a result, indoor thermal comfort has an immense influence on the occupants' mental and physical health, performance, productivity, and decision-making capabilities \cite{[75]}. 
This is particularly true for primary school students, who spend a considerable amount of time in classrooms (4553 hours, on average) \cite{OECD}. There is enough evidence to demonstrate that satisfactory thermal comfort in classrooms enhances students' concentration levels and improves learning outcomes \cite{[53]}.

 Thermal comfort (TC) perception depends on measurable parameters (indoor temperature, relative humidity etc) and personal factors (clothing insulation, activity levels etc). It is also equally subjective and depends on individual preference. Thus, estimating and predicting the TC of an occupant is a non-trivial task.
 Conventional TC estimation solutions viz., Predicted Mean Vote, Predicted Percentage of Dissatisfied, and Adaptive models do not offer accurate TC prediction \cite{ML_TC_REVIEW_1}. Consequently, researchers have turned to 
 machine learning (ML) for thermal comfort prediction. ML-based models have frequently been shown to be more robust and reliable than conventional solutions~\cite{ML_TC_REVIEW_1, ML_TC_REVIEW_2}. 
More importantly, ML-based data-driven models are capable of optimizing energy consumption of smart 
HVAC systems \cite{yang2020a}. However, the energy cost of HVACs is very high, comprising almost 50\% of the energy consumption of a building \cite{hvacenergy}. 

Thus, naturally ventilated (NV) buildings are ideal from the perspective of energy efficiency and sustainability \cite{ventilation2}. 
Further, in most developing countries, the bulk of indoor spaces, including classrooms are naturally ventilated. 
\color{black}
However, in an NV building, the indoor thermal conditions are unregulated, leaving the occupants vulnerable to the external environment. 
This makes estimating and predicting occupants' thermal comfort in NV buildings a complex problem. Therefore, while naturally ventilated buildings are perfect for sustainable cities, there is a need for smart human-centered prediction systems that maintain satisfactory levels of thermal comfort.

\subsection{Motivation and Research Problems}
The indoor thermal environment of an NV building is not controlled by an HVAC system. Thus, the characteristics of an NV building such as its location, design, layout, cross-ventilation strategies, construction materials and insulation directly influence occupant thermal comfort. 
This makes thermal comfort in NV buildings a space-dependent phenomenon, thereby causing spatial variability in the performance of TC prediction models in different buildings. From the perspective of practical application, it is pertinent to investigate the variability in the performance of TC prediction models across NV buildings. 
Further, it is essential to understand which features determine the model performance in a building-specific spatial context. The generalization capability of models is also of extreme importance, as it indicates the feasibility of training a model once and applying it to different spatial contexts.

A major inspiration for this work is designing human-centric prediction models based on real-world data. To that end, naturally ventilated schools make a suitable site for field experiments and data gathering. They are a key component in energy conservation and combating climate change \cite{ventilation3}. Moreover, studies on NV classrooms and young children are lacking in the fast-expanding body of ML-based TC prediction research \cite{ML_TC_REVIEW_1, ML_TC_REVIEW_2}. The focus has mainly been on building ML models for adult occupants that optimize indoor thermal comfort through efficient control of HVAC and air-conditioning systems. Also, a comprehensive dataset for primary school students' TC analysis is unavailable in the standard \mbox{ASHRAE-I} and ASHRAE-II databases \cite{db1, db2}. Moreover, finding TC prediction solutions for primary school students is important from the perspective of technological applications for societal welfare. Young students have limited cognitive abilities to evaluate the classroom environment, leading to data bias and imbalance~\cite{bettyDeepcomfort}.
Consequently, predicting their thermal comfort status in uncontrolled NV spaces adds greater complexity to the underlying ML classification problems, as compared to adults. 
\begin{figure}[h]
\centering 
    \includegraphics[width=\linewidth]{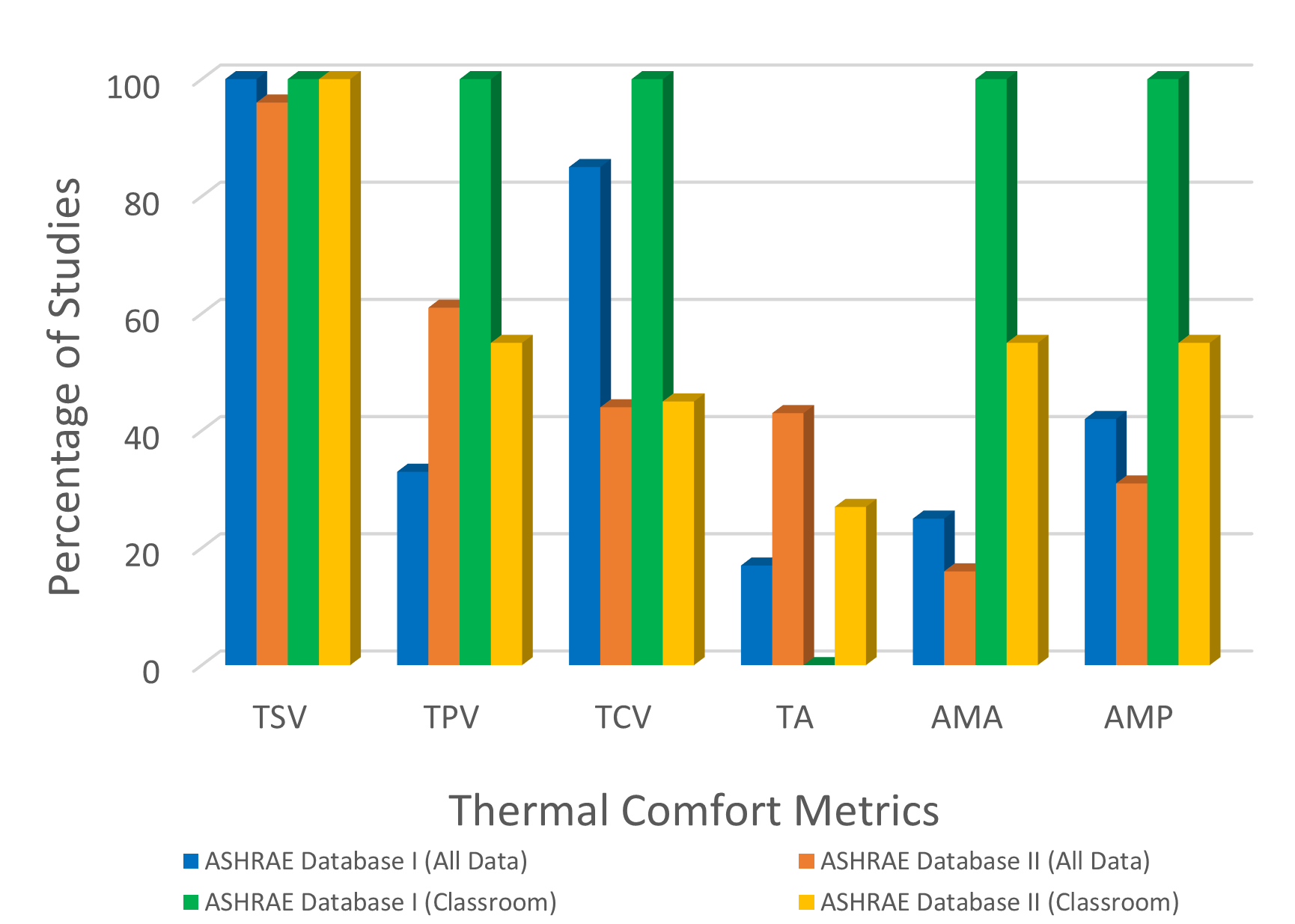}
\caption{TC Metrics in ASHRAE Databases I \& II}
\label{fig:AshTCmetrics}
\vspace{-0.5cm}
\end{figure}

\begin{figure*}[h]
    \centering
    \vspace{0pt}
    \begin{minipage}[t][][b]{0.60\textwidth}
     \centering%
\begin{tabular}{cc}
    \subfloat {\includegraphics[width=0.33\linewidth]{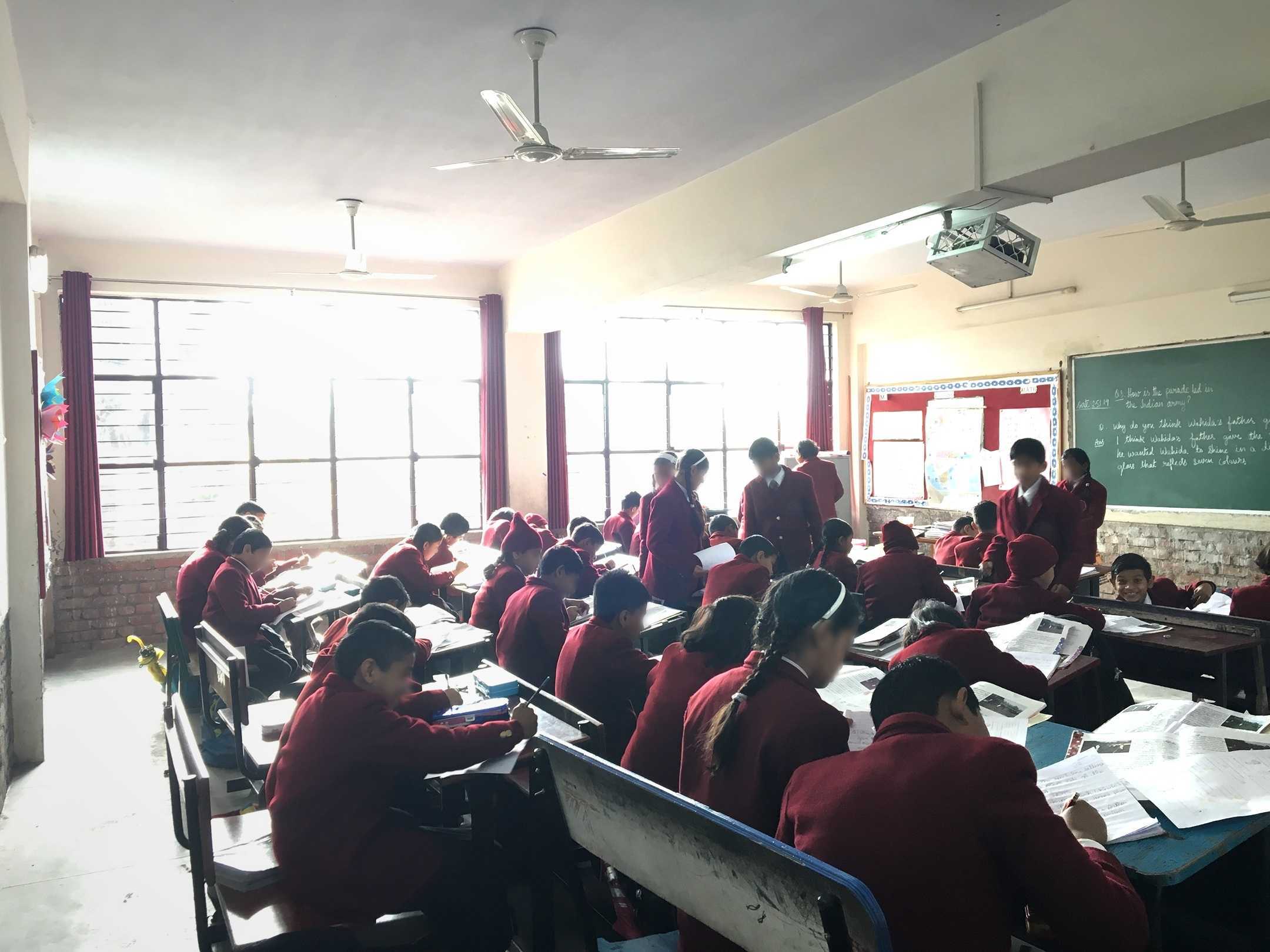}}\hfill\hspace{0.1cm}%
\subfloat{\includegraphics[width=0.33\linewidth]{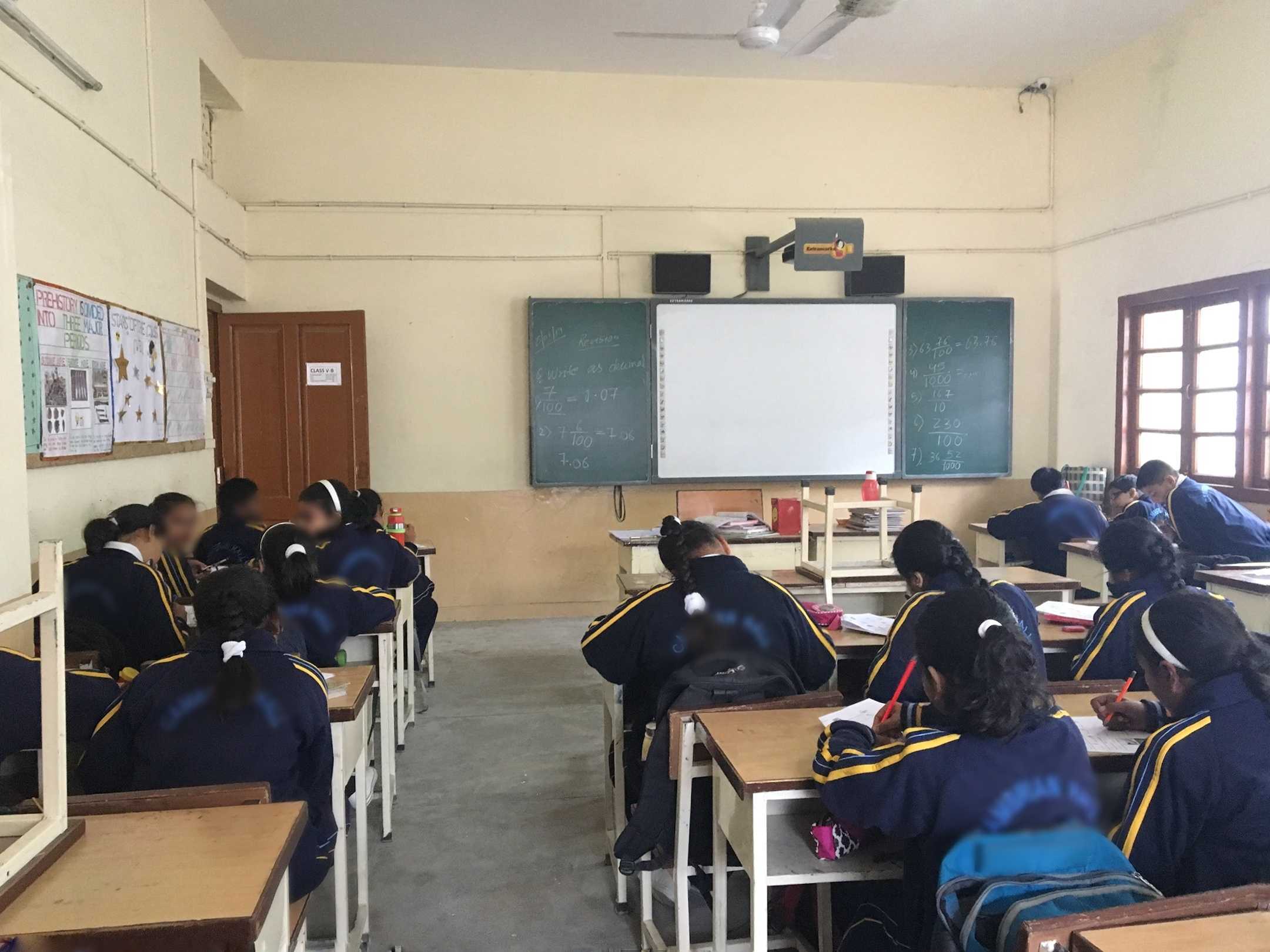}}\hfill\hspace{0.1cm}
  \subfloat{\includegraphics[width=.33\linewidth]{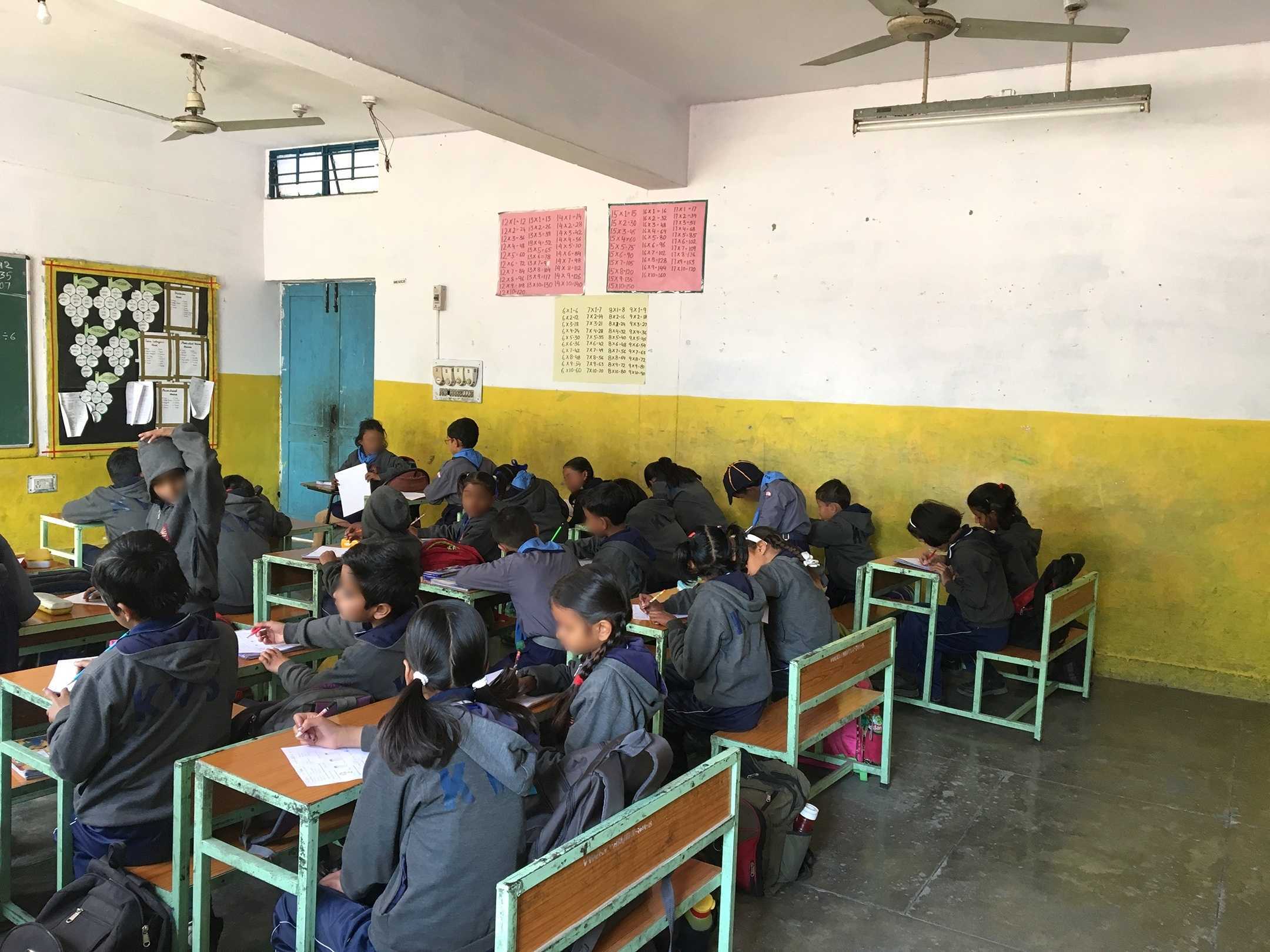}}\hfill\hspace{0.1cm}%
\end{tabular}
      \caption{Field Experiments in Naturally Ventilated Classrooms} 
          \label{fig:photos}
    \end{minipage}
    \vspace{0pt}
     \begin{minipage}[t][][b]{0.3\textwidth} 
\scriptsize
  \label{tab:example}
  \centering
  \captionof{table}{Statistical Details}
    \label{surveystats}
  \begin{tblr}{
      colspec={cp{0.5cm}p{0.75cm}p{0.5cm}},
      row{1}={font=\bfseries},  
      column{1}={font=\itshape},
      row{even}={bg=gray!10},
    }
    \toprule
             School & Student Grades  & No. of Days  & Unique Students \\
    \toprule
    1&3,4,5&5&103\\
    2&3,4    &4&74\\
    3&3,4,5&4&135\\
    4&3,4,5&5&82\\
    5&3,4,5&2&118
    \bottomrule
  \end{tblr}
  
    \end{minipage}
\end{figure*}

\subsection{Contributions}

Having discussed the main research problems this work aims to solve, the major contributions are outlined below:
 
 \begin{itemize}
     \item Field Experiments and Surveys: Month-long field experiments were conducted in 14 classrooms of 5 NV schools involving 512 unique participants (students).
     \item Data analysis: Distribution analysis of subjective TC responses is performed to investigate the variation across school buildings.
     \item Predictive Modeling: ML-based prediction of 3 primary Thermal Comfort metrics. TC Model performance is evaluated with respect to: (a) Spatial Variability across the 5 Schools (b) Generalization capability of the models.
     \item Spatial context of feature importance: Variation of feature importance across the 5 buildings is analyzed. 
     \color{black}
     \item Comparative Analysis with ASHRAE-II: Prediction models for the primary students dataset are compared with TC models for adults utilizing ASHRAE-II dataset.
  \end{itemize}
It is worth highlighting that the ASHRAE-II database~\cite{db2} does not include a dataset for naturally ventilated classrooms of India, and the dataset created in this work will bridge that gap \footnote{Data and code used in this study will be temporarily made available here: https://bit.ly/3yXUJkR. The combined winter and summer data along with the code will be publicly released through the lab's website.}. 


\section{ML-Based Thermal Comfort Prediction} \label{sec:review}
This section presents a brief review of ML-based TC prediction (MLTC) studies relevant to this work. Starting with the scope, ML models can be designed and trained to predict individual thermal comfort \cite{han2020a} or group-based comfort model (GCM) \cite{chai2020a}. GCMs are more scalable and feasible to implement in public spaces such as a classroom. However, achieving high prediction accuracy in a GCM is difficult. This work implements GCMs for students of a school as a whole. 
Further, a variety of features are considered in TC prediction, which include indoor and outdoor environmental parameters, physiological and demographic data, and other individual parameters \cite{ML_TC_REVIEW_1, ML_TC_REVIEW_2}. The primary student dataset in this work is created through indoor and outdoor field-experiments, student questionnaires, and weather data procured from Indian Meteorological Department (IMD) for the month of the study. Features include classroom environment parameters (e.g., indoor temperature), individual parameters (e.g., Clothing value, metabolic rate), demographic parameters (e.g., age, gender), building related parameters (e.g., building type, ventilation type, etc.), weather data, and miscellaneous categorical features (e.g., survey timing and school type).  

Coming to objectives, ML models generally predict subjective comfort responses of occupants. 
Figure~\ref{fig:AshTCmetrics} shows frequently used thermal comfort metrics which serve as model outputs/labels viz., Thermal Sensation Vote (TSV), Thermal Preference Vote (TPV), and Thermal Comfort Vote (TCV). Less popular metrics include Thermal Acceptability (TA), Air Movement Acceptability (AMA) and Air Movement Preference (AMP). 
The majority of ML-based thermal comfort (MLTC) research focuses on adult participants aged 20--30s \cite{ML_TC_REVIEW_1}. However, primary school children are not only more vulnerable to an unfavorable thermal environment due to their limited adaptive capacity, they are also less likely to express their discomfort due to psycho-social constraints~\cite{[65], bettyDeepcomfort}. 
These factors introduce additional complexity in predicting their thermal comfort perceptions~\cite{bettyDeepcomfort}. Recent literature reviews highlight the lack of adequate research on ML-based thermal comfort prediction with children as the primary participants \cite{ML_TC_REVIEW_1, ML_TC_REVIEW_2}. 

On a positive note, new TC studies have begun to apply advanced ML techniques such as \textit{multi-task learning} to predict children's thermal comfort~\cite{bettyDeepcomfort}. However, aspects such as spatial and temporal variability, and their impact on the TC perception of school going children remains unexplored. This work bridges the gap by focusing on spatial variability. The next section describes the field-experiments and the data gathering exercise.



\begin{figure}[h]
\centering 
    \includegraphics[width=\linewidth]{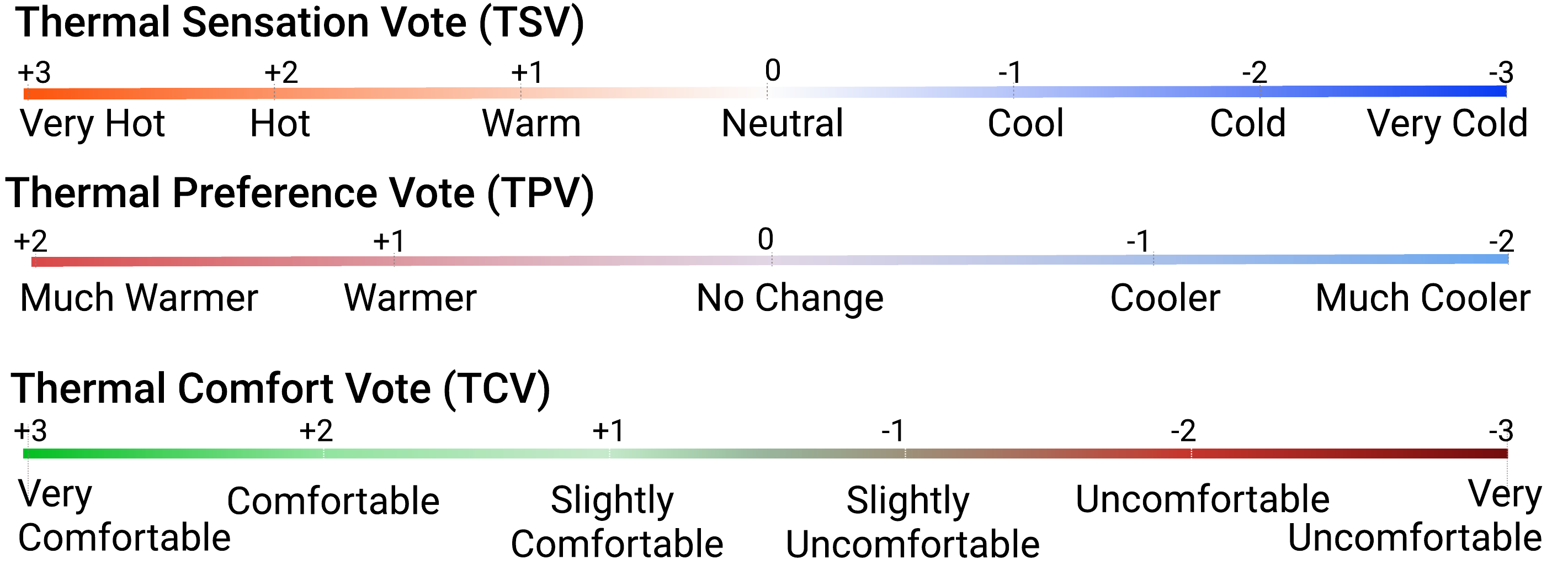}
\caption{Thermal Comfort Metric Scales with Values}
\label{fig:tcmetrics}
\end{figure}

\section{Field Experiments and Data Gathering} \label{sec:survey}

The field experiments were conducted in 5 reputed schools in Dehradun city,  known as the ``School capital of India,” viz., Grace Academy, St Thomas School, Kendriya Vidhyalaya, Cambrian Hall and Jaswant Model School. 
To ensure confidentiality, the gathered data is anonymized. 
Likewise, in the discussion ahead, the schools are represented as School$_{i}$, where $i\in \{1\ldots5\}$ is randomly allotted to a school.
The primary school students from class-levels/grade-levels 3$^{rd}$ to 5$^{th}$, typically belonging to ages 6-13, are the participants of this study. A few classrooms where the field experiments were conducted are presented in Figure~\ref{fig:photos}.

\begin{figure}[h]
\captionof{table}{IoT Sensors used in Measurements}
\includegraphics[width=\linewidth]{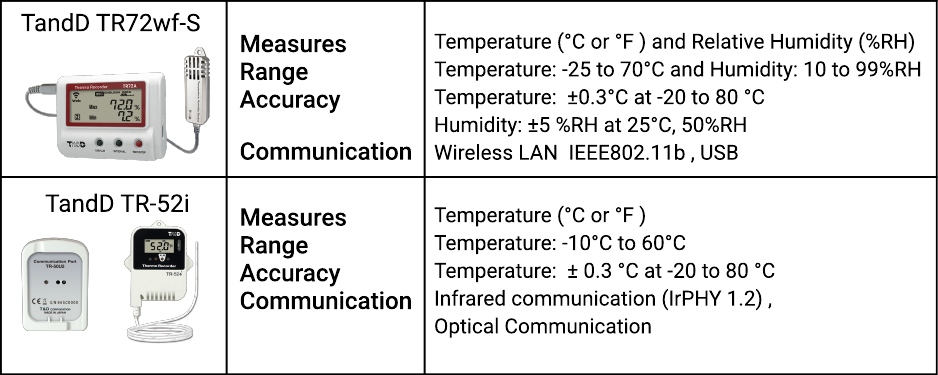}
\label{table:iot}
\end{figure}
The dataset comprises of 2039 responses collected from 512 primary school children as unique participants. A school-wise quantitative distribution of participants is presented in Table~\ref{surveystats}. The study was conducted in January, the coldest winter month in the region. To capture the maximal impact of cold on student comfort, field experiments were conducted between 8:30 AM--12 PM (divided into 6 half-hour time slots), and 93\% samples were collected before 11 AM. An illustration-rich questionnaire was specifically designed for young students with limited comprehension and cognitive abilities. A total of 21 subjective (e.g., comfort level) and objective (e.g., clothes worn) questions were asked. Students assessed their indoor thermal comfort on six TC metrics.
The multi-point scales of the three TC metrics considered in this work, viz., TSV, TPV, and TCV, 
are depicted in Figure~\ref{fig:tcmetrics}. 
\color{black}
Two IoT sensors were used to gather indoor temperature, outdoor temperature, and indoor relative humidity. Their details are presented in Table~\ref{table:iot}. Detailed descriptions of the methodology, field experiments, and questionnaire survey are available in~\cite{bettyDeepcomfort}.

\begin{figure*}[ht!]
 \centering%
\begin{tabular}{cc}
  \subfloat[School Locations] {\includegraphics[width=.19\linewidth]{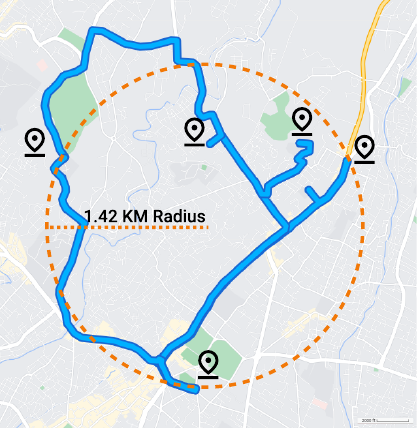}}\hspace{0.1cm}%
    \subfloat[TSV] {\includegraphics[width=.27\linewidth]{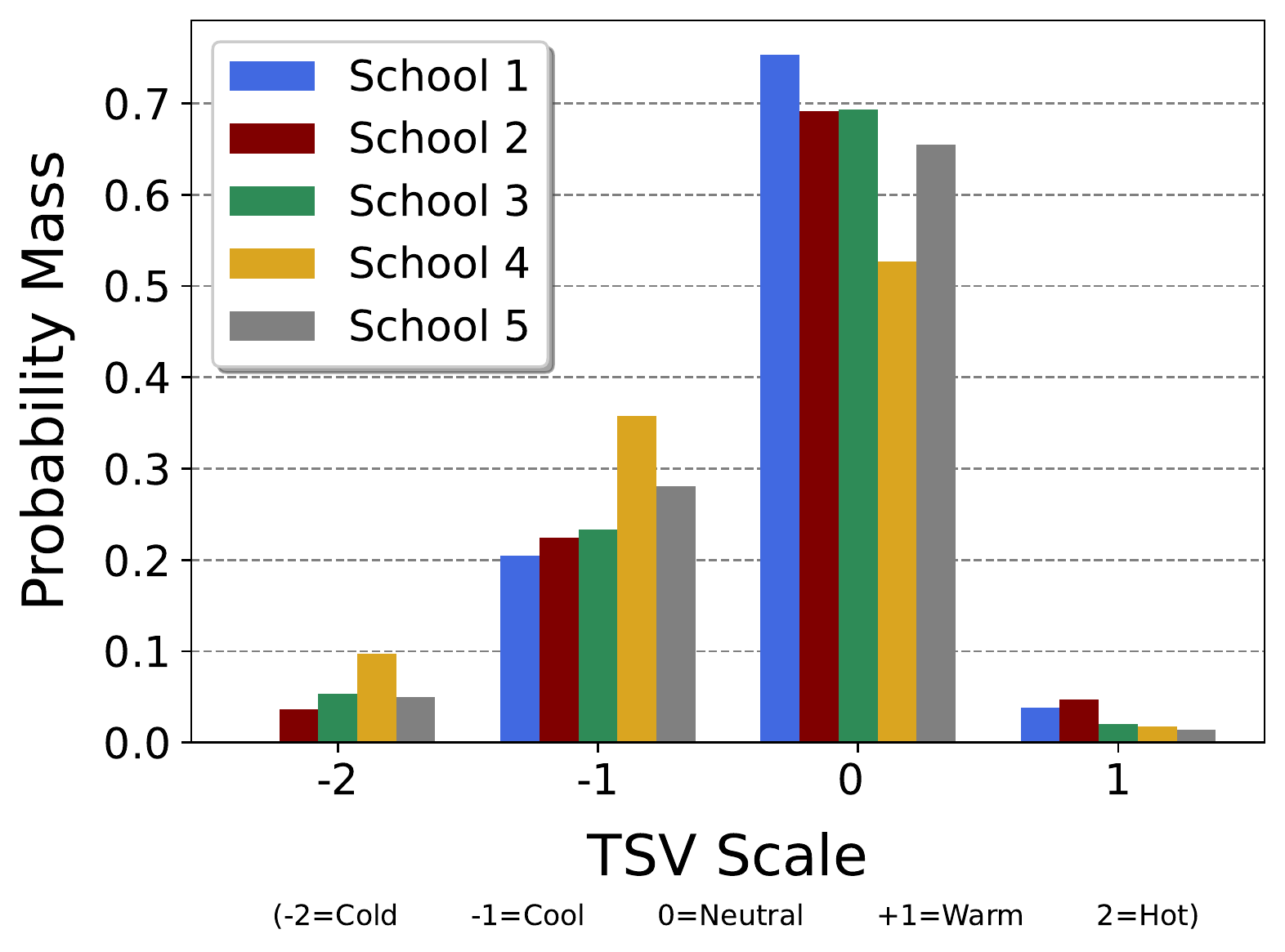}}\hfill%
\subfloat[TPV] {\includegraphics[width=.27\linewidth]{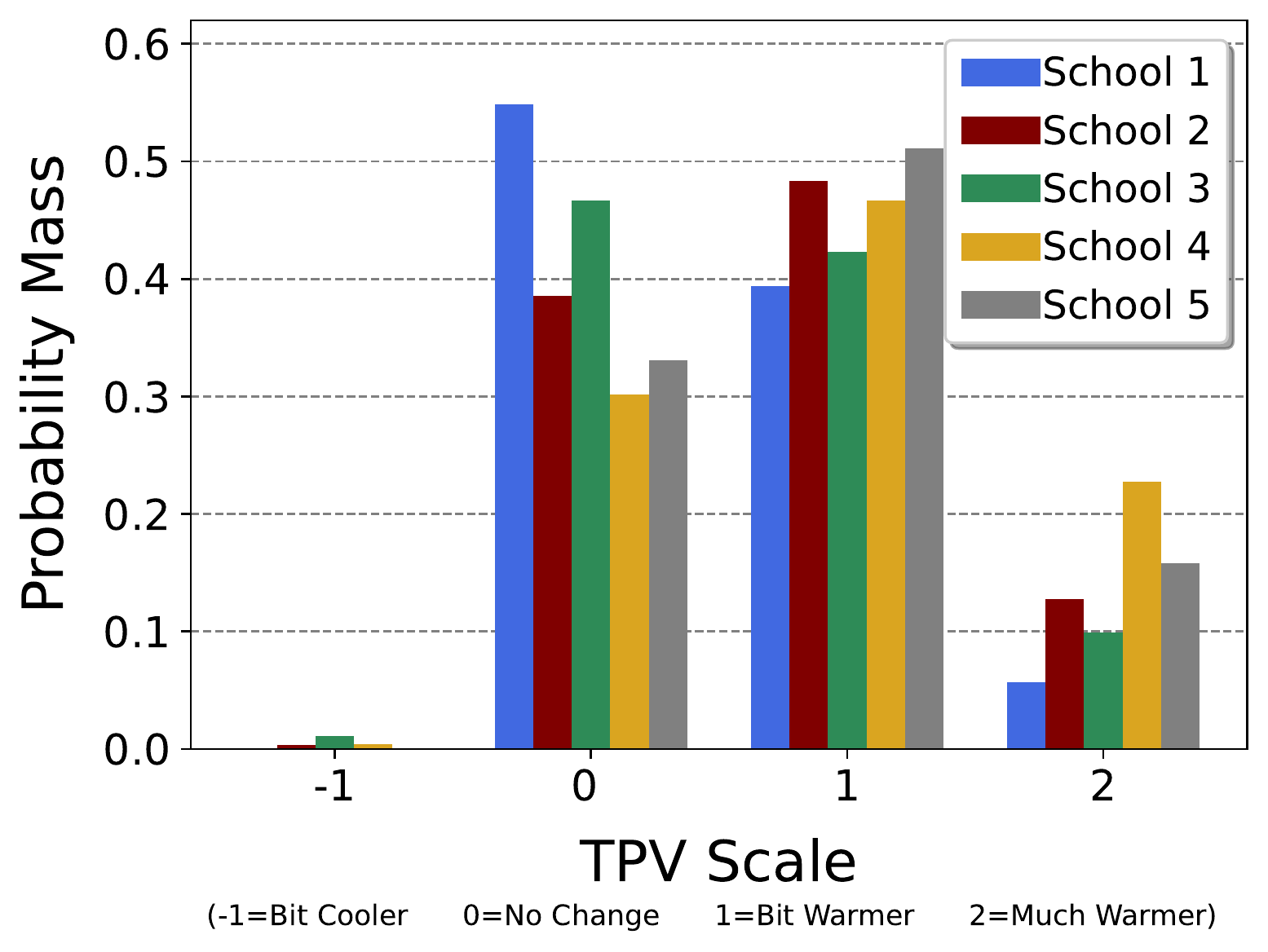}}\hfill
  \subfloat[TCV] {\includegraphics[width=.27\linewidth]{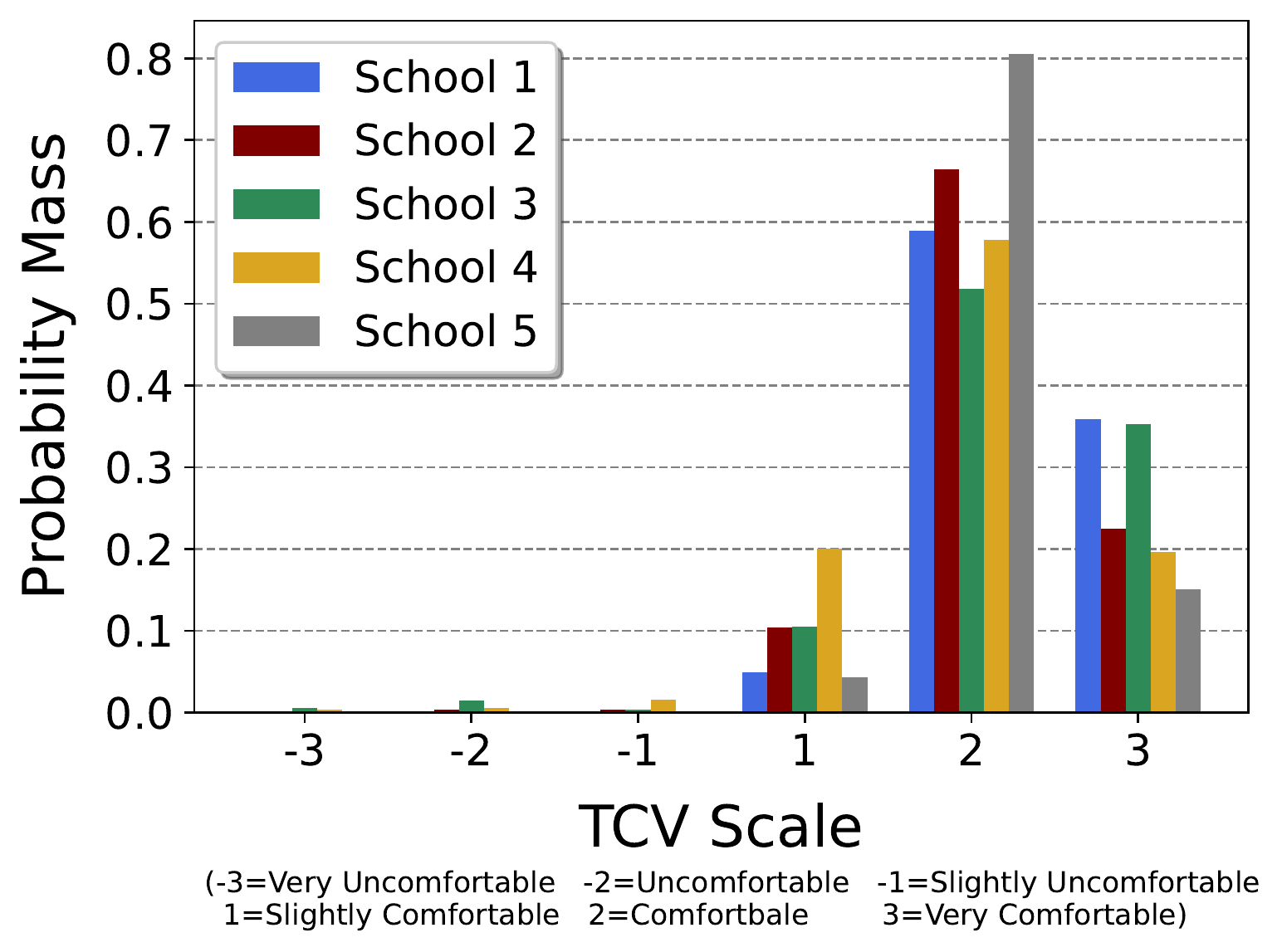}}\hfill%
\end{tabular}
      \caption{School Locations and Probability Distribution of Output Metrics}  
    \label{fig:pdfSchools}
\end{figure*}

\section{Exploratory Data Analysis}\label{sec:dataanalysis}
  The 5 schools lie within a radius of 1.42 Km (Figure~\ref{fig:pdfSchools}(a)), with similar levels of vegetation and elevation, and are all naturally ventilated. 
  Further, the prescribed ``school uniforms'' also have similar \textit{Clothing} values. 
  
  Despite these similarities, the spatial environments in the 5 schools show marked differences. 
 The inspection of school-sites and buildings revealed architectural variation across the five schools. The schools differ in terms of the number of floors, window sizes, structure, facade, building layout, and orientation. Further, the number and orientation of windows and the size of classrooms varied as well. Building walls and windows were not fully insulated in all schools. In some classrooms, the ventilators (higher window openings in the corridor-side wall) were closed in winters as a protective measure against cold, but it also affected the cross-ventilation. Moreover, elements of campus planning, e.g., open courtyard plan for improved cross-ventilation, were noticed in the architectural layout of three out of the five schools. 
 
 Due to the difference in these factors, variation in parameters such as indoor temperature and relative humidity were recorded across schools and even across classrooms in the same school. Though they are in the same city, the indoor environmental quality differed considerably across schools. Thus, it is relevant to study how the TC perception of students varies across the 5 schools and how it affects prediction performance.

  The distribution of TSV, TPV, and TCV responses is presented in Figure~\ref{fig:pdfSchools}(b), Figure~\ref{fig:pdfSchools}(c), and Figure~\ref{fig:pdfSchools}(d), respectively.
  It is discernible that in School~1 the highest proportion of primary students have responded to feeling ``Neutral'' (TSV=0) and prefer ``No Change" to their classroom environments (TPV=0). In sharp contrast, School~4 has the highest proportion of students who feel ``Cold'' or ``Cool'' (TSV=-2,-1). 
 The typical causes of discomfort to students in classrooms include
 cold/hot wind draft, vertical air temperature difference, radiant asymmetry, etc. \cite{almeida2016thermal}.
 While the effect of local discomfort on students’ well being and performance is well documented, determining the precise impact of each local spatial factor, is a complex problem.
\begin{figure}[h]
\centering 
    \includegraphics[width=\linewidth]{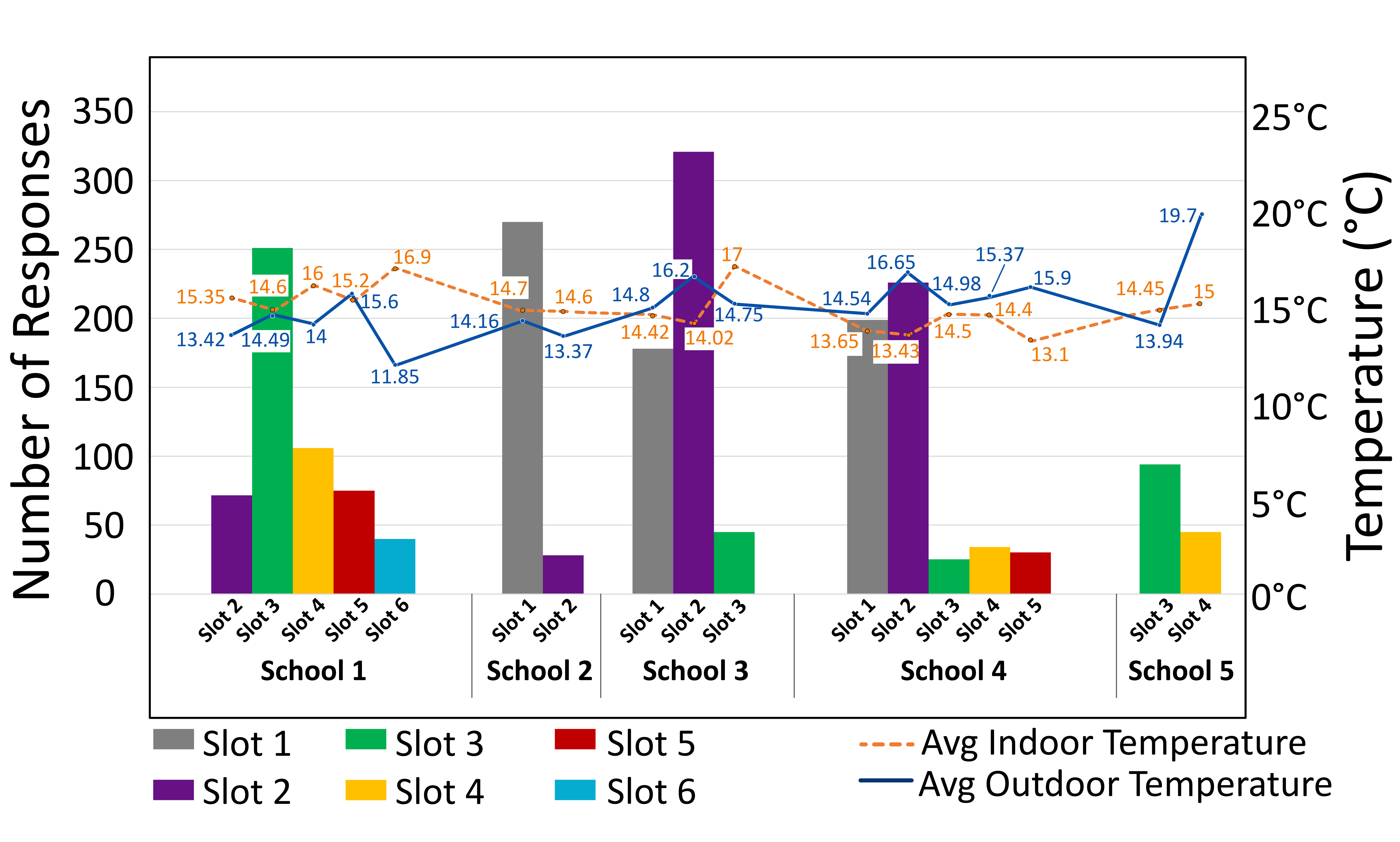}
\caption{Time-slots, Indoor, \& Outdoor Temperatures}
\label{Schoolfactors}
\end{figure}
 For example, it was observed that an opening in the passageway with no shutters in a classroom of School~4 allowed a constant draft of cold wind. Further, while two classrooms in School~1 were exposed to direct sun, the classrooms in School~4 did not receive any sunlight during the survey time. 

However, the distribution of comfort response votes in Figure~\ref{fig:pdfSchools} is also influenced by other factors. 
This is visible in the data presented in Figure~\ref{Schoolfactors}. The day of the experiment, the field experiment timings, and the number of responses, vary across schools. Further, the classroom temperatures and outdoor temperatures during the experiments also differ. Differences in the average indoor and outdoor temperatures for each slot in each school, presented in Figure~\ref{Schoolfactors}, varies from 0.17°C to 3.2°C. 
It is ensured that all of these factors (features) are used to train the TC prediction models so that their impact is adequately reflected. In the generalization ability test, it is ensured that a trained model only tests samples from experiment slots it has been trained on, as the indoor and outdoor temperatures vary with time. Further, the impact of buildings on TC perception and prediction is ascertained through spatial variability in \textit{feature importance}. Variation in features that are related to the naturally ventilated built environment will demonstrate the important role buildings play.
 

\begin{figure*}[htbp]
 \centering%
\begin{tabular}{cc}
  \subfloat[Student vs Adults] {\includegraphics[width=.17\linewidth]{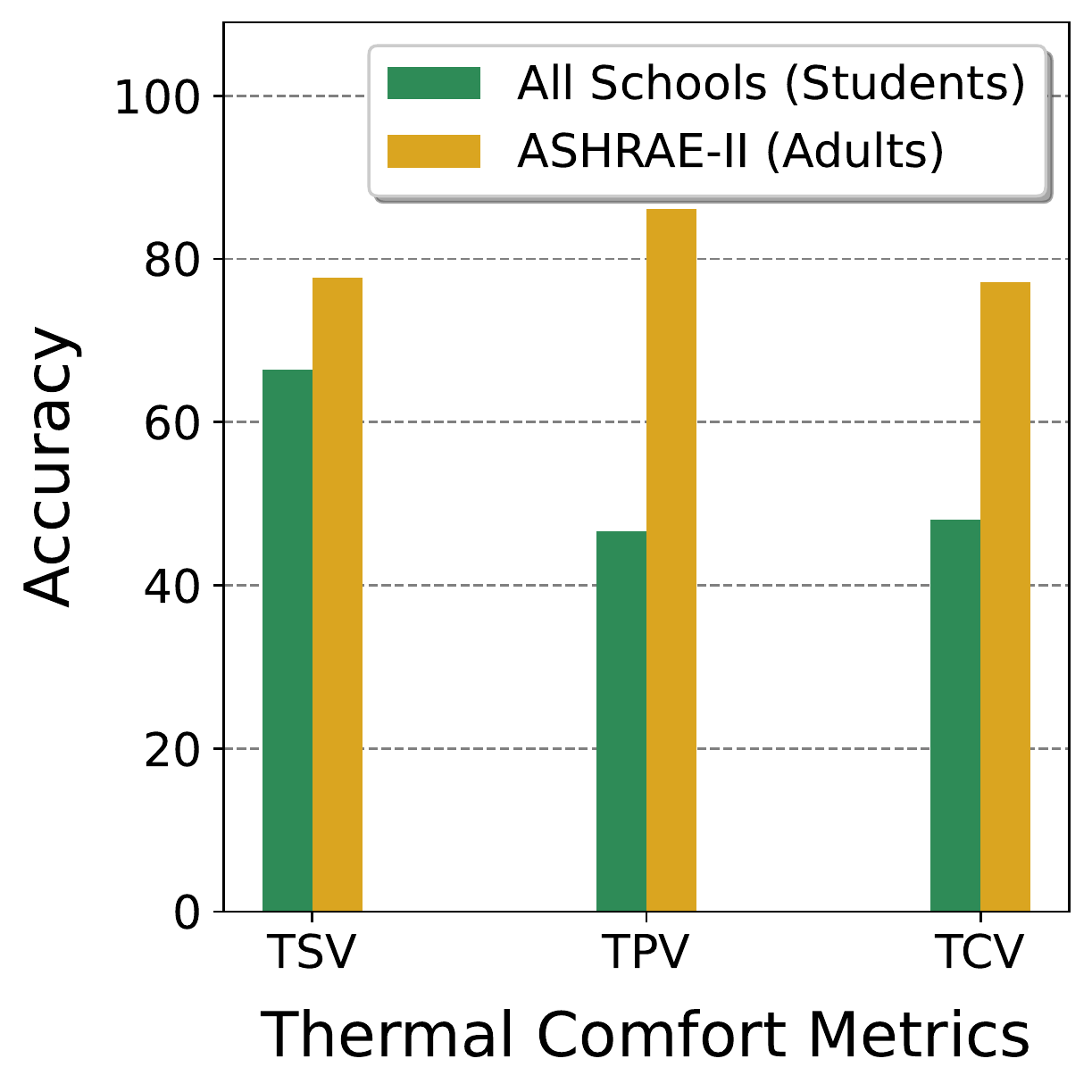}}
  \subfloat[School 1] {\includegraphics[width=.17\linewidth]{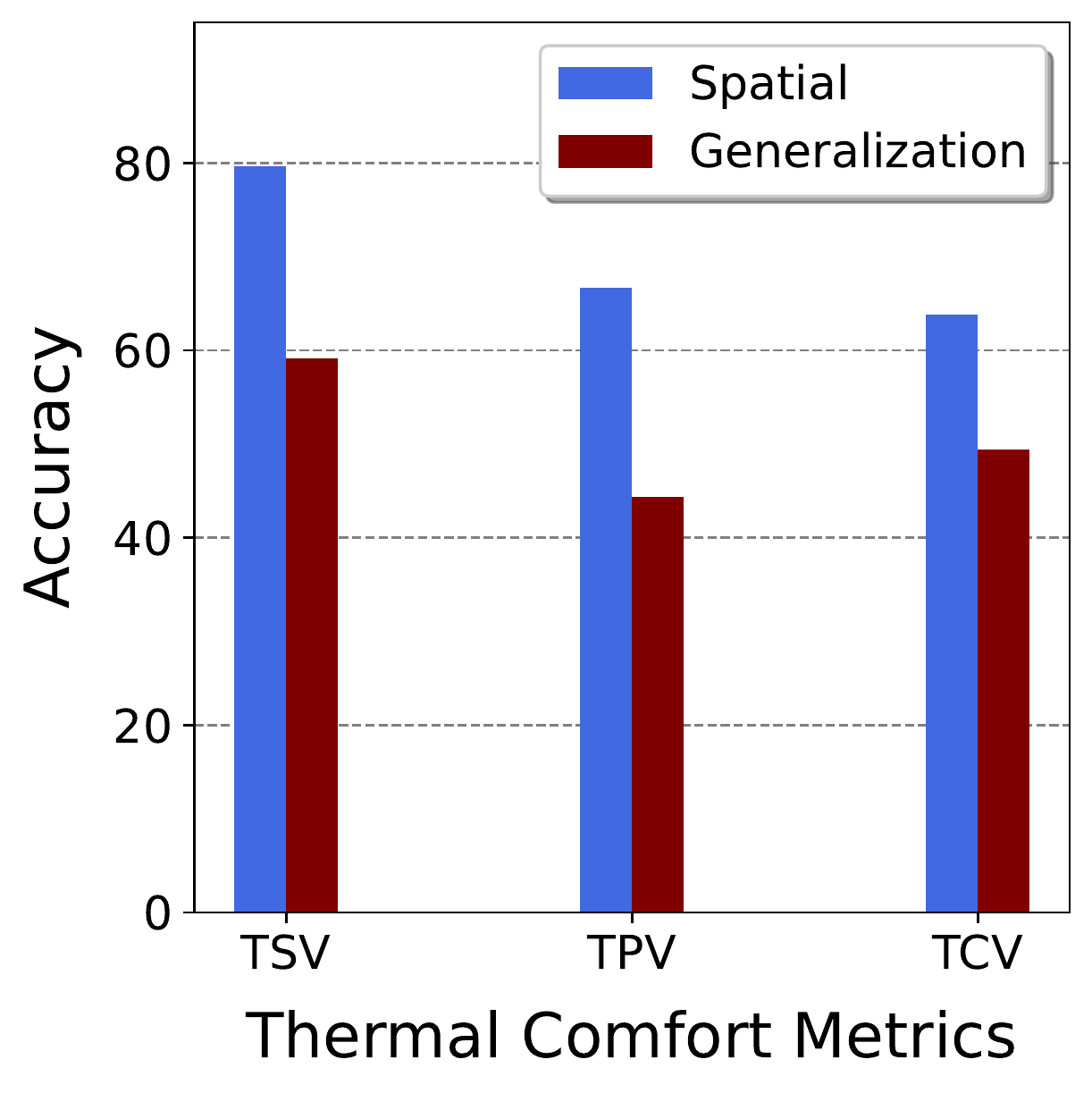}}
  \subfloat[School 2] {\includegraphics[width=.17\linewidth]{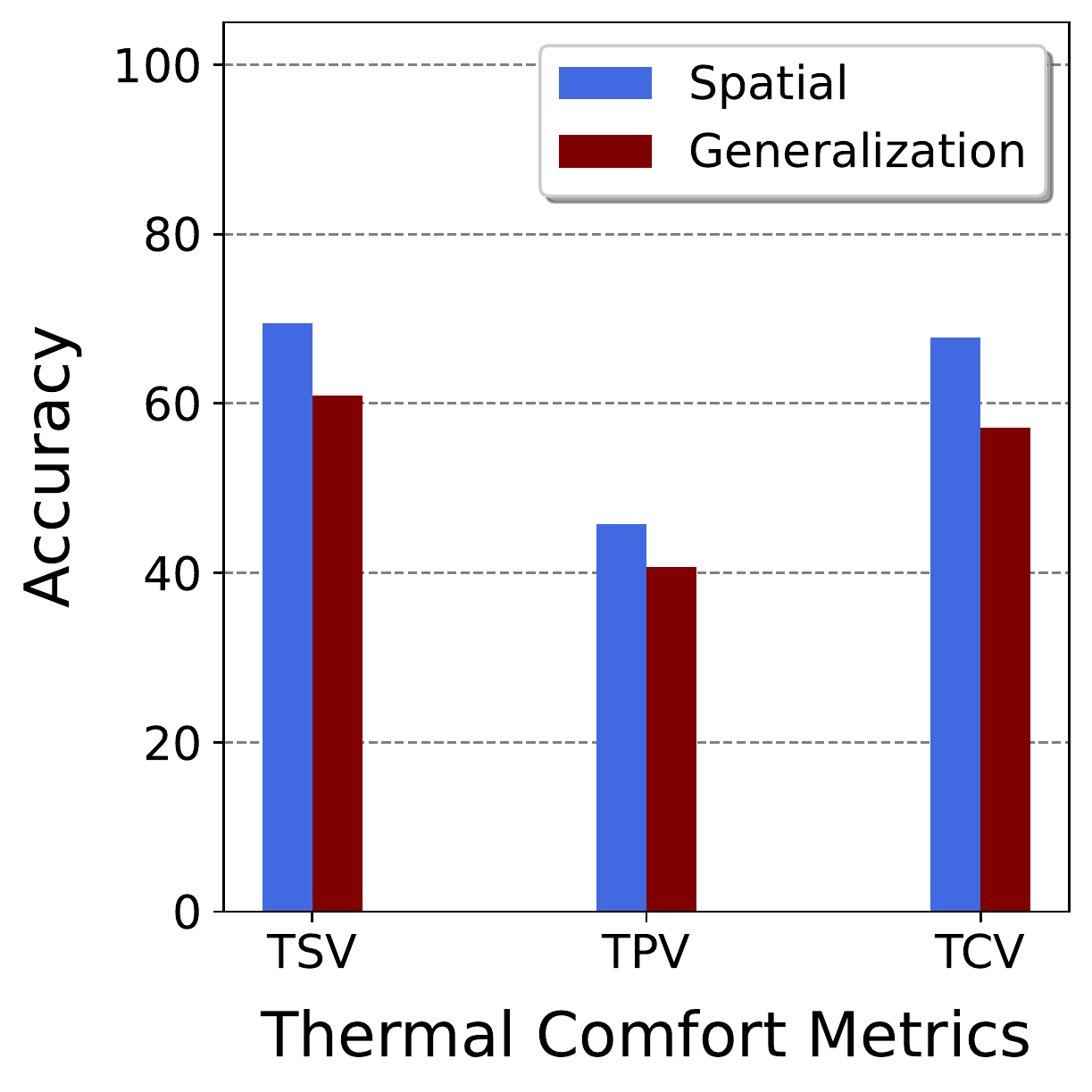}}
  \subfloat[School 3] {\includegraphics[width=.17\linewidth]{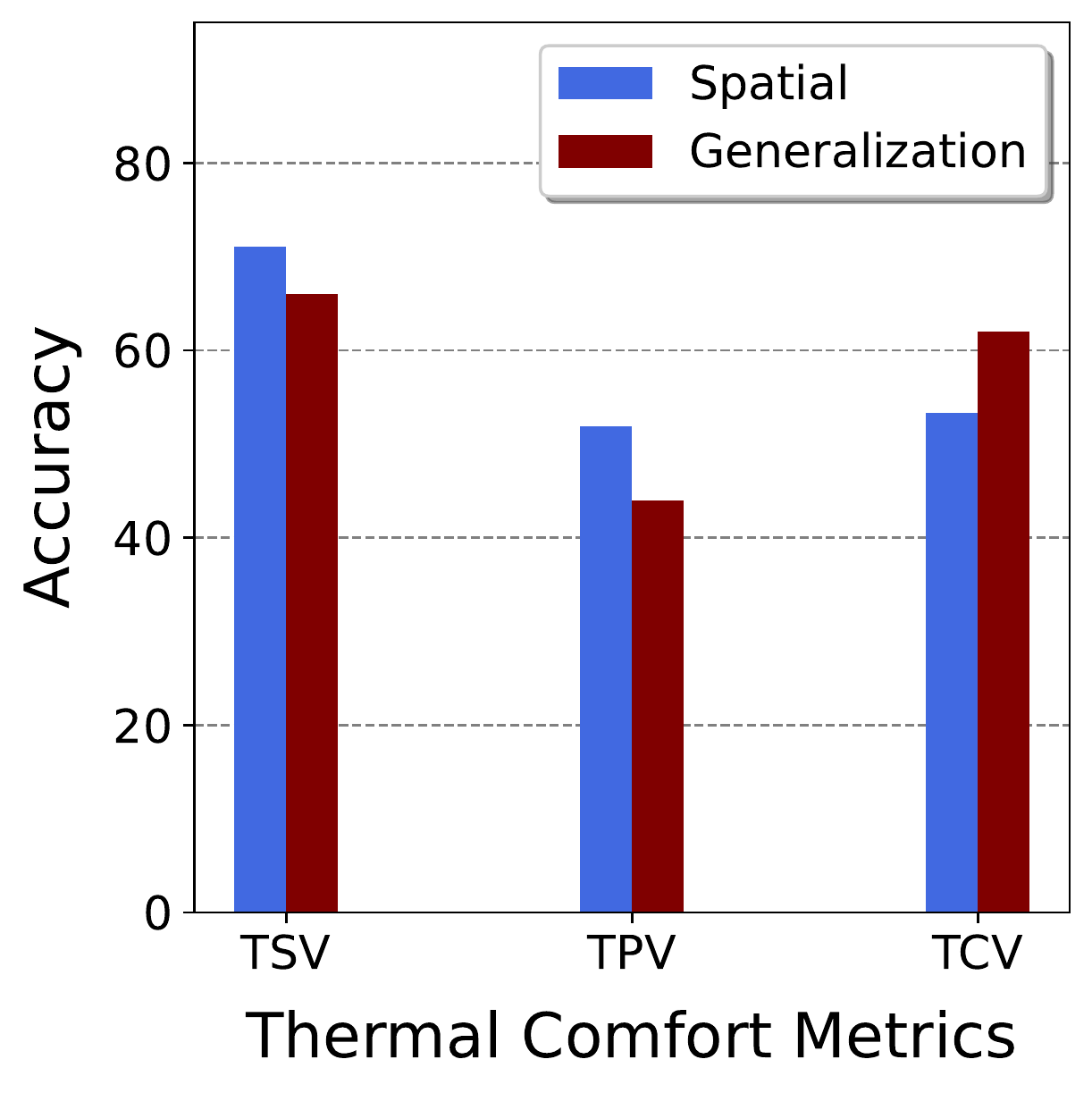}}
  \subfloat[School 4] {\includegraphics[width=.17\linewidth]{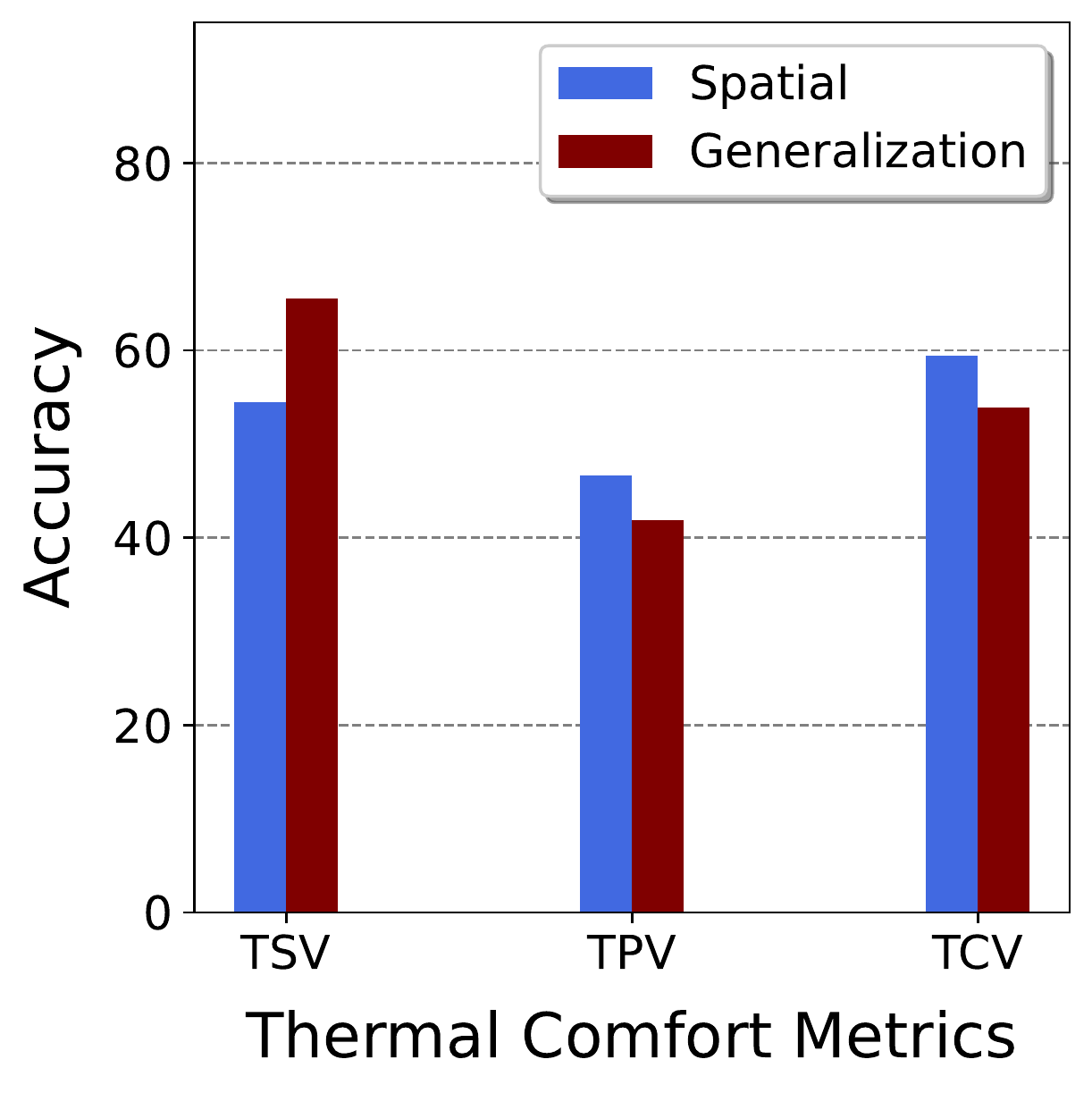}}
  \subfloat[School 5] {\includegraphics[width=.17\linewidth]{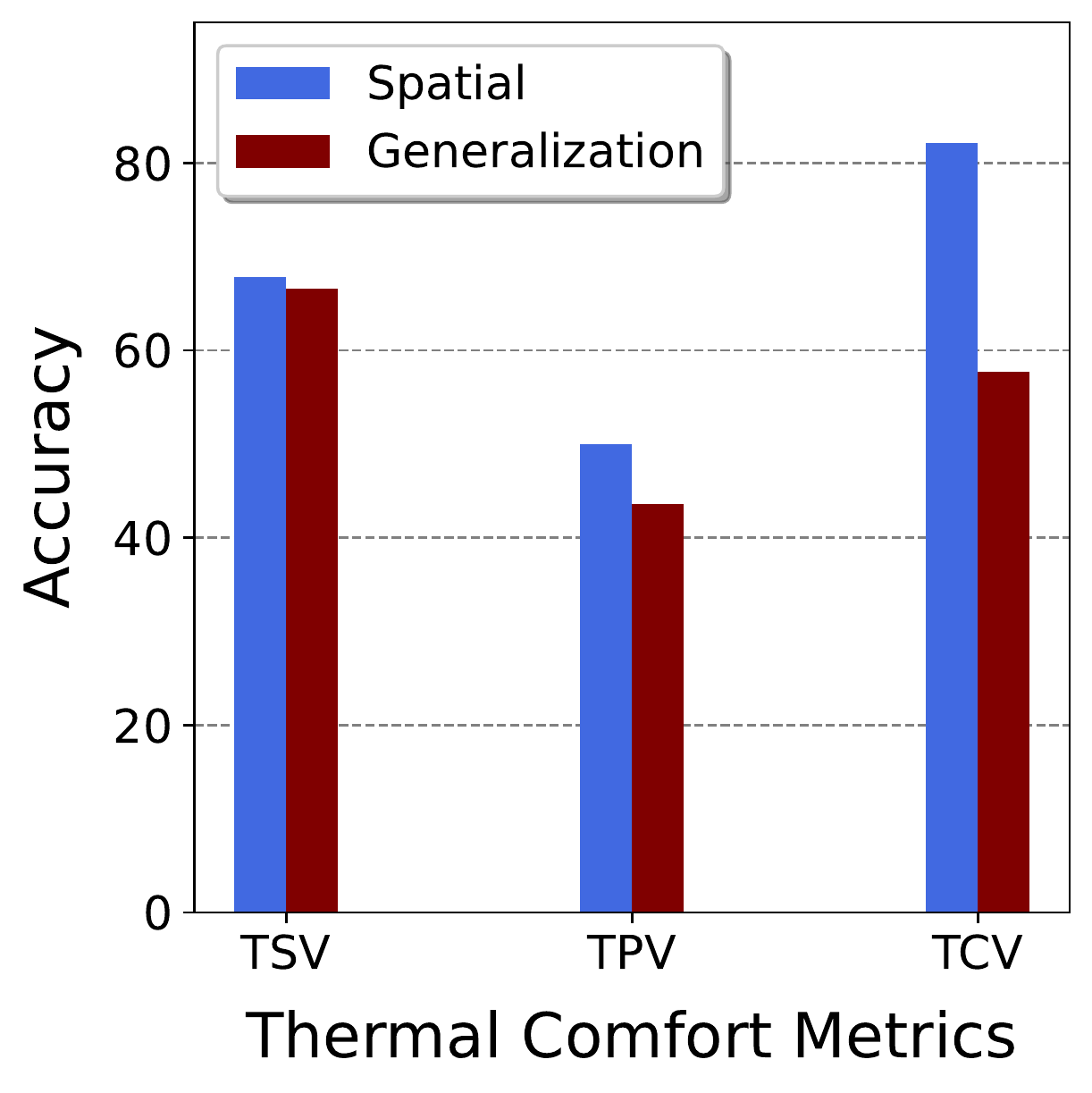}}

\end{tabular}
      \caption{Accuracy of SVM prediction models}  
    \label{fig:results}
\end{figure*}
\par\textbf{Challenges in Multi-class Classification:}
A comparative analysis of the distribution of TSV, TPV, and TCV responses in 
Figure~\ref{fig:pdfSchools} 
reveals the complexity in TC prediction for primary school students.
A smaller percentage of students responded to feeling a ``Cool'' or ``Cold'' sensation (TSV=-1 or -2), while a much larger proportion of student responses indicate that they prefer the classroom environment to be a ``Bit Warmer'' or ``Much Warmer'' (TPV=1 or 2). What makes the problem more challenging is that
most students 
claim to be experiencing varying degrees of comfort (TCV=1 or 2 or 3), which contradicts the TSV and TPV trends. 
The presence of these ``illogical votes,'' in data adversely impacts the Accuracy, Precision, and Recall of multi-class classification models for primary students, which is demonstrated ahead.
\color{black}

\section{Predictive Modeling of Thermal Comfort}~\label{sec:implementation}
\textbf{Problem Formulation:}
Let $\mathcal{D}: \{(\v x_i, \v y_i)\}_{i=1}^{N}$ be the dataset where, $N$ is number of points, $\mathcal{C}$ is number of classes. Moreover, $\v x_i \in \mathbb{R}^d$ is the feature vector, and $\v y_i \in \{0, 1\}^{\mathcal{C}}$ is the label vector for $i^{th}$ point. 
The model and evaluation strategy are explained below.

\par\textbf{Support Vector Machine~(SVM):} 
For the given dataset, the SVM model attempts to solve the following optimization problem for binary classification:
\begin{align} \label{svm_eqn}
    \min_{\v w, b} \frac{1}{2} \v w^{\top}\v w + C \sum_{i=1}^{N} {\zeta_i} \\  \mbox{s.t.      } y_i(\v w^{\top} \phi(\v x_i) + b) &\ge 1 - \zeta_i, \forall i \nonumber
    \zeta_i &\ge 0, \forall i \nonumber
\end{align}
where, $y_i$ is +1 if the label is assigned to $i^{th}$ instance and -1 otherwise. Here $\v w \in \mathbb{R}^d$ represents the weight-vector and $b$ represents the bias. $C$ is a hyperparameter set through cross-validation which controls the importance of L2 regularization term. The dual of Eq.~\ref{svm_eqn} can be written as:
\begin{align} \label{svm_dual}
    \max_{\v \alpha} \sum_{i=1}^{N} \v \alpha_i - \frac{1}{2} \sum_{i,j=1}^{N} y_i y_j \alpha_i  \alpha_j \phi(\v x_i)^{\top} \phi(\v x_j)\\  \mbox{s.t.      } 0 \le \v \alpha_i \ge C, \ \forall i\nonumber\\ \sum_{i=1}^{N} \alpha_i y_i = 0 \nonumber
\end{align}
Eq.~\ref{svm_dual} is solved using the libsvm algorithm~\cite{libsvm}. Note that regularization may avoid overfitting, thus helping the classifier to generalize better on unseen samples. Given that TC metric prediction may involve a non-linear fit, the kernel function $\phi$, facilitates the learning of non-linear class-boundaries. The kernel function allows us to compute the dot product between two vectors ($\v u$, $\v v$) in arbitrarily large spaces ($\phi(\v u)^{\top} \phi(\v v)$) without explicitly projecting the vectors into high dimensional space ($\phi(\v u)$ or $\phi(\v v)$). Results are reported for multiple kernels, viz., Linear, Polynomial kernels of degrees 2 and 3, 
and Radial Bias Function (RBF).

\par\textbf{Objectives:} The aim is to ascertain: (i) The impact of \textit{spatial variability} on classification model performance for individual schools: data for each individual school was considered during training and a school-specific test performance is reported, (ii) The \textit{generalization ability} of the models across schools: $(n-1)$ schools were considered during training and test performance is reported on the $n^{th}$ school, and (iii) Spatial context of feature importance: SVM models were trained and tested on each individual school's data and feature importance is reported. 

There is an inherent class imbalance (Figure~\ref{fig:pdfSchools}) in the data which needs to be kept in account during training as well as evaluation. For the spatial variation test-scenario that involves data from each individual school, label values with extreme class imbalance ($\leq3\%$ of total sample size) were not considered. For the other two test-scenarios, i.e., investigation of generalization ability and feature importance, no data pruning was performed and the entire dataset was considered. To overcome the class imbalance, the training pipeline assigns a weigh to each class, where, $\beta_{+} = \frac{N}{\sum_{i=1}^{N}\mathbb{I}(y_{i}=1)}$ and $\beta_{-} = \frac{N}{\sum_{i=1}^{N}\mathbb{I}(y_{i}=-1)}$ are the weights for the positive and negative class, respectively. The SVM optimization problem in ~\ref{svm_eqn} incorporates the class weights by reformulating $C = \beta C$. Further, the train-test set were created using stratified sampling which preserves the class distribution. 

Two additional points are noteworthy. First, thermal comfort prediction is a multi-class classification problem. Thus, the \textit{one-vs-rest} strategy is used to convert the binary classification that can be solved using Eq.~\ref{svm_eqn} into a multi-class classification problem. Secondly, in the spatial variability analysis, ML models are trained and tested on data from individual schools, which reduces the overall sample space. Thus, SVM is used to avoid overfitting as it is a suitable classification algorithm when the sample space is not large. Other probable solutions include data augmentation and employing domain specific constraints, which are beyond the scope of this work.

\begin{table*}[h]
     \centering
    \caption{Spatial Variability in TC Prediction}
    \label{tab:spatial}
    \begin{tabular} {cccccccccc}
      \toprule
      \multirow{4}{*}{School} &
        \multicolumn{9}{c}{Thermal Comfort Output Metrics}\\
        \cmidrule(lr){2-10}
        &
        \multicolumn{3}{v{15em}}{Thermal Sensation Vote} &
        \multicolumn{3}{v{15em}}{Thermal Preference Vote} &
        \multicolumn{3}{v{15em}}{Thermal Comfort Vote} \\
        \cmidrule(lr){2-4}\cmidrule(lr){5-7}\cmidrule(lr){8-10}
        &
         \multicolumn{1}{v{4.5em}}{F1-score} &
        \multicolumn{1}{v{4.5em}}{Precision} &
        \multicolumn{1}{v{4.5em}}{Recall} &

         \multicolumn{1}{v{4.5em}}{F1-score} &
        \multicolumn{1}{v{4.5em}}{Precision} &
        \multicolumn{1}{v{4.5em}}{Recall} &
         \multicolumn{1}{v{4.5em}}{F1-score} &
        \multicolumn{1}{v{4.5em}}{Precision} &
        \multicolumn{1}{v{4.5em}}{Recall}\\
      \cmidrule(l){1-1}\cmidrule(l){2-2}\cmidrule(l){3-3}\cmidrule(l){4-4}%
        \cmidrule(l){5-5}\cmidrule(l){6-6}\cmidrule(l){7-7}\cmidrule(l){8-8}\cmidrule(l){9-9}\cmidrule(l){10-10}
School 1&71&84&80&64&72&67&62&60&64\\
School 2&57&48&69&47&59&46&55&46&68\\
School 3&59&50&71&53&59&52&37&28&53\\
School 4&41&57&54&30&22&47&44&35&59\\
School 5&55&46&68&33&25&50&74&67&82\\
All Schools&53&	44&	66&	48&	50&	47&	52&	59&	48  \\
ASHARE-II\textsuperscript{*}&78&80&78&86&87&86&77&79&77\\
      \bottomrule
     \end{tabular}\vfill
     {\raggedright{\tiny \textsuperscript{*}Adult data for naturally ventilated Buildings was considered.}}
\end{table*}
\section{Results and Analysis}~\label{sec:results}
The SVM models were trained on our primary student dataset and the ASHRAE-II dataset and the results for the three objectives are presented below.

\subsection{SVM Performance on Adult vs. Primary Student Data}
SVM models are trained on ASHRAE-II dataset for adults with a similar context as our primary student data, i.e., data for naturally ventilated buildings in winters. As shown in Figure~\ref{fig:results}(a), SVM performance on ASHRAE-II data exceeds its prediction Accuracy on ``All Schools" data by 16.9\%, 85.0\%, and 60.5\% for TSV, TPV, and TCV, respectively. Likewise, with respect to F1-scores, the predictions for ASHRAE-II adult data outperform primary student predictions by 46.73\%, 81.3\%, and 48.4\%, respectively. 
It is interesting to note that students are able to assess their thermal sensation (TSV) more accurately than their thermal preference (TPV) or thermal comfort levels (TCV). 
This finding highlights the complexity involved in predicting primary students' indoor comfort. Their inability to accurately evaluate their environment and express their level of comfort can be attributed to limited cognition and reasoning, high agreeableness, and lack of opportunities to modify the classroom environment. Adults do not seem to face such difficulties. 

\subsection{Spatial Variability of Thermal Comfort Prediction}
Results presented in Figure~\ref{fig:results} and Table~\ref{tab:spatial} demonstrate that the indoor spaces of NV buildings have an immense influence on thermal comfort prediction. Models trained and tested on data of specific schools lead to higher prediction performance in general, when compared to models trained and tested on data of all schools. This is visible from Accuracy of ``All Schools" in Figure~\ref{fig:results}(a) and those of individual schools in  Figures~\ref{fig:results}(b)--\ref{fig:results}(f). However, the magnitude of difference fluctuates across TC metrics. For example, the increase in Accuracy upon considering the spatial context, for TSV, TPV, and TCV, lies in the range of 2.16\%--19.85\%, 7.36\%--43.16\%, 10.88\%--71\%, respectively. The individual school performance is lower than ``All Schools" in only two instances out of fifteen, viz., TSV of School~4 has and TPV of School~2.   
In School~4, the spatial context is challenging for TC prediction as discussed in Section~\ref{sec:dataanalysis}. Inclusion of additional features such as air velocity and illuminance can perhaps improve the prediction performance. The decrease in Accuracy in case of School~2  is marginal (1.73\%).
Thus, spatial context does have a huge impact on the prediction performance of subjective TC metrics, especially ``comfort levels".

This finding is further reinforced by the variation in all performance measures, viz., Accuracy, F1-score, Precision, and Recall, across the different schools. The standard deviation (Std Dev) in Accuracy for TSV, TPV, and TCV for the 5 schools is 9.06, 8.43, and 10.85, respectively. Likewise, for F1-score, the Std Dev is 10.89, 14.10, and 14.50, respectively. Further, the magnitude of the maximum difference between schools in the Accuracy of TSV, TPV, and TCV prediction is 46.18\%, 45.69\%, and 54.19\%, respectively. Please note that features such as comprehension of students, age, and gender will typically vary across schools and contribute to the overall performance difference. Nevertheless, the high degree of spatial variation in thermal comfort perception and prediction in naturally ventilated buildings is evident. 

In the school-specific asessment, it is particularly interesting to observe that for children, prediction performance of ``sensation" (TSV) and ``comfort level" (TCV) models is better than those of ``preference" (TPV) models. However, the reverse is true for adults, whose TPV model's performance is better than that of TSV and TCV models~(Figure~\ref{fig:results}(a)). A logical inference can be made that children find it difficult to express their ``preference." 
As highlighted earlier, expressing thermal sensation seems to be the easiest as it involves a highly objective assessment.
\begin{table*}[h]
    \centering
\centering
    \caption{Generalization Ability of SVM TC Models}
    \label{tab:general}
    \begin{tabular}{cccccccccc}
      \toprule
      \multirow{4}{*}{School} &
        \multicolumn{9}{c}{Thermal Comfort Output Metrics}\\
        \cmidrule(lr){2-10}
        &
        \multicolumn{3}{v{15em}}{Thermal Sensation Vote} &
        \multicolumn{3}{v{15em}}{Thermal Preference Vote} &
        \multicolumn{3}{v{15em}}{Thermal Comfort Vote} \\
        \cmidrule(lr){2-4}\cmidrule(lr){5-7}\cmidrule(lr){8-10}
        &
    
         \multicolumn{1}{v{4.5em}}{F1-score} &
        \multicolumn{1}{v{4.5em}}{Precision} &
        \multicolumn{1}{v{4.5em}}{Recall} &
         \multicolumn{1}{v{4.5em}}{F1-score} &
        \multicolumn{1}{v{4.5em}}{Precision} &
        \multicolumn{1}{v{4.5em}}{Recall} &
         \multicolumn{1}{v{4.5em}}{F1-score} &
        \multicolumn{1}{v{4.5em}}{Precision} &
        \multicolumn{1}{v{4.5em}}{Recall}\\
      \cmidrule(l){1-1}\cmidrule(l){2-2}\cmidrule(l){3-3}\cmidrule(l){4-4}%
        \cmidrule(l){5-5}\cmidrule(l){6-6}\cmidrule(l){7-7}\cmidrule(l){8-8}\cmidrule(l){9-9}\cmidrule(l){10-10}
School 1	&	53&	50&	59&	38&	41&	44&	45&	43&	49\\
School 2	&	53&	53&	61&	42&	44&	41&	43&	40&	57\\
School 3	&	54&	52&	60&	41&	42&	44&	47&	38&	62\\
School 4	&	59&	51&	71&	42&	43&	42&	49&	48&	54\\
School 5	&	53&	61&	67&	27&	19&	44&	42&	33&	58\\
\bottomrule
\end{tabular}
\end{table*}

\subsection{Generalization Capability of SVM Models}
Generalization capability of TC models is desirable from the perspective of \textit{train-once-deploy-anywhere} paradigm. However, given the spatial and temporal contexts of thermal comfort and the subjectivity of individual TC perception, high generalization ability is difficult to achieve. This is even more true for NV buildings and primary students. The first step towards achieving generalization across spatio-temporal contexts is to test the  generalization ability of popular low-cost ML algorithms such as SVM. 

Thus, SVM models were trained on $n-1$ schools, and tested on the $n^{th}$ school, and the results are presented in Figures~\ref{fig:results}(b)--\ref{fig:results}(f) and Table~\ref{tab:general}. Three trends in the results stand out. First, the generalization ability of SVM based TC models is not very high but quite stable (low variation). For example, the Accuracy of TSV models has a median of 65.3, and a low Std Dev of 3.37. Second, the spatial school-specific models outperform generalization models in all but two test-scenarios, viz., TCV of School~3 and TSV of School~4. However, apart from School~1, the difference in prediction performance is not very huge.
Finally, similar to the school-specific analysis, generalization models seem to be better at predicting ``sensation" (TSV) and ``comfort" (TCV) as compared to thermal ``preference" (TPV). 

An important conclusion that can be drawn from the generalization ability analysis is that train-once-deploy-anywhere models may perform well for similar type of spatial contexts, viz., naturally ventilated buildings. Further, the low variation in model performance also suggests the suitability of advanced ML techniques such as transfer learning.

\begin{figure}[h]
\centering 
    \includegraphics[width=\linewidth]{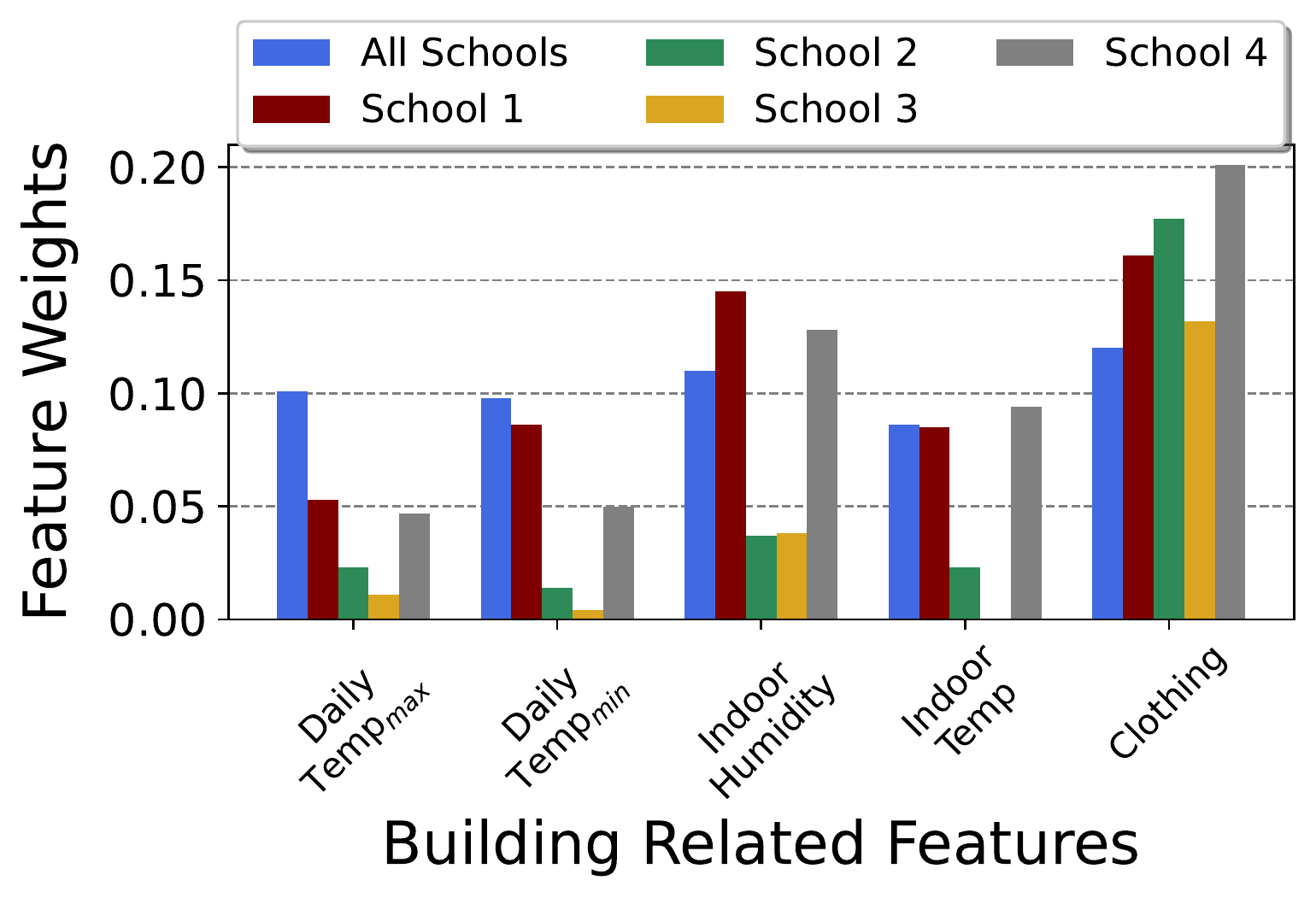}
\caption{Spatial Context of Feature Importance}
\label{fig:featureimp}
\end{figure}
\subsection{Feature Importance and Building Environments}
\color{black}The \textit{Permutation Feature Importance} technique is used to determine the relative importance of the features. This technique is independent of the ML algorithm and is used as a standard to compare the relative importance of features used for prediction. 
It fits a model on the train dataset, and then makes predictions on the test dataset where the values of the features are scrambled. Finally, the mean importance scores are generated by making predictions on the scrambled features. Although the technique is algorithm agnostic, there may be some variance depending on the stochastic nature of the algorithm.


\color{black}Figure~\ref{fig:featureimp} illustrates the variation in feature importance in model performance for the schools. School~5 is not considered due to its low number of samples. It can be observed that the selected features have a strong relationship with the building environment, i.e., they either influence it (e.g., maximum daily outdoor temperature) or are influenced by it (e.g., clothing values). The high variability in feature importance demonstrates that the roles that these features play varies drastically across the 5 naturally ventilated school buildings. It is correct that there is some variation across schools in other features as well. However, the high variation in the importance of building related features, indicates that model performance has a strong spatial context. 

\section{Major Takeaways and Way Forward}~\label{sec:conclusions}
The primary hypothesis of this work, that thermal comfort in naturally ventilated buildings has a strong spatial component, stands validated. The findings have the following major implications: (1) For high accuracy, TC models need to be trained with highly spatial or building-specific data. Despite this, performance may vary across naturally ventilated buildings. (2) TC models can be generalized across spaces for real world use-cases, viz., train on one NV building, and deploy the model on neighboring NV buildings. Techniques such as SVM offer high stability in prediction accuracy, but for practical applications it needs to be improved. New paradigms such as Transfer Learning and Multi-task Learning (3) Children are able to ``sense'' thermal discomfort but unable to express it clearly in terms of ``preference'' or ``comfort levels.'' Thus, the twin challenges of illogical responses and class imbalance need to be addressed.

The next steps involve extending the scope of this work to field-experiments in the summer in more schools, employing a wider set of ML algorithms including deep learning models, and evaluating prediction performance in hyper-local contexts  such as classrooms. We have also found strong evidence of temporal variability of TC prediction, and intend to explore it through a time-of-day analysis. 

\section{Acknowledgment}
The research was funded by the Sasakawa Scientific Research Grant of the Japan Science Society and JSPS KAKENHI Grant Number JP 22H01652. Authors thank the principals, teachers, and students of the 5 schools in Dehradun city, India, that participated in the study. Authors also thank Mrs. Pushpa Manas, Director of School Education (Retd.), Uttarakhand, India, for facilitating this study.

\bibliography{ref, MLTCR1, MLTCR2}
\end{document}